\documentclass[letterpaper, 10 pt, conference]{ieeeconf}
\IEEEoverridecommandlockouts                              

\usepackage{pdfpages}
\usepackage{amsmath}
\usepackage{graphics} 
\usepackage{epsfig} 
\usepackage{mathptmx} 
\usepackage{times} 
\usepackage{amsmath} 
\usepackage{dirtytalk}
\usepackage{amssymb}  
\usepackage{color}
\usepackage{caption}
\usepackage{subcaption}
\usepackage{diagbox}
\usepackage{algorithm}
\usepackage{algpseudocode}
\usepackage{makecell}
\usepackage{array}
\usepackage{svg}

\usepackage{booktabs}
\title{\LARGE \bf
DriveGenVLM: Real-world Video Generation for Vision Language Model based Autonomous Driving
}

\author{Yongjie Fu, Anmol Jain, Xuan Di$^{*}$, 
Xu Chen, and Zhaobin Mo
\thanks{$^{*}$\textit{Corresponding author: Xuan Di.}}
\thanks{$^{\text{‡}}$This work is sponsored by NSF CPS-2038984 and NSF ERC-2133516.}
\thanks{Yongjie Fu is with the Department of Civil Engineering and Engineering Mechanics,
        Columbia University, New York, NY, 10027, USA (E-mail: yf2578@columbia.edu).}
\thanks{Anmol Jain is with the Department of Computer Science,
        Columbia University, New York, NY, 10027, USA (E-mail: aj3231@columbia.edu).}
\thanks{Xuan Di is with the Department of Civil Engineering and Engineering Mechanics, Columbia University, New York, NY, 10027 USA, and also with the Data Science Institute, Columbia University, New York, NY, 10027 USA (E-mail: sharon.di@columbia.edu).}
\thanks{Xu Chen and Zhaobin Mo are with the Department of Civil Engineering and Engineering Mechanics,
        Columbia University, New York, NY, 10027, USA (E-mail: xc2412@columbia.edu, zm2302@columbia.edu).}
\thanks{$**$ This work has been accepted by 2024 IEEE International Automated Vehicle Validation Conference (IAVVC 2024)}
}

\begin{document}

\maketitle
\thispagestyle{empty}
\pagestyle{empty}

\begin{abstract}
The advancement of autonomous driving technologies necessitates increasingly sophisticated methods for understanding and predicting real-world scenarios. Vision language models (VLMs) are emerging as revolutionary tools with significant potential to influence autonomous driving. In this paper, we propose the DriveGenVLM framework to generate driving videos and use VLMs to understand them. To achieve this, we employ a video generation framework grounded in denoising diffusion probabilistic models (DDPM) aimed at predicting real-world video sequences. We then explore the adequacy of our generated videos for use in VLMs by employing a pre-trained model known as Efficient In-context Learning on Egocentric Videos (EILEV). The diffusion model is trained with the Waymo open dataset and evaluated using the Fréchet Video Distance (FVD) score to ensure the quality and realism of the generated videos. Corresponding narrations are provided by EILEV for these generated videos, which may be beneficial in the autonomous driving domain. These narrations can enhance traffic scene understanding, aid in navigation, and improve planning capabilities. The integration of video generation with VLMs in the DriveGenVLM framework represents a significant step forward in leveraging advanced AI models to address complex challenges in autonomous driving.

\end{abstract}

\section{INTRODUCTION}
\label{sec-intro}

In the rapidly evolving field of autonomous driving, the integration of advanced predictive models into vehicular systems or transportation systems has become increasingly critical for enhancing safety and efficiency \cite{fu2024digital, fu2023federated}. Among the myriad of sensory technologies employed, camera-based video prediction stands out as a pivotal component, offering a dynamic and rich source of real-world data. Through the adoption of a cutting-edge diffusion model approach, this research not only contributes to the advancement of autonomous driving technologies but also sets a new benchmark for the application of predictive models in enhancing vehicular safety and navigational precision.

Content generated by artificial intelligence is presently a leading area of study within the domains of computer vision and artificial intelligence. The generation of photo-realistic and coherent videos is one of the challenging areas because of the limitations of memory and computation time. In the autonomous vehicle area, predicting the video from a vehicle's front camera is crucial for several reasons, particularly in the context of autonomous driving and advanced driver-assistance systems (ADAS) \cite{ohn2015surveillance}. In this paper, we utilize the videos from the vehicle's surrounding cameras and predict future frames.  

The generative model has also been utilized in the area of transportation and autonomous driving \cite{mo2022trafficflowgan, mo2022quantifying}. Models are increasingly recognized for their capability to understand driving environments. Vision language models (VLMs) are now being utilized for autonomous driving applications. To enhance the utility of VLMs and explore the application of generative models to video content within VLMs, it is essential to validate generative models' predictions to confirm their relevance and accuracy in real-world scenarios. DriveGenVLM introduces the in-context VLM as a method to validate predicted videos from a diffusion-based generative model by providing textual descriptions of driving scenarios.

\subsection{Related Work}

Diffusion-based architectures have become increasingly popular in recent research for generating images and videos. Diffusion models have been applied to a variety of tasks for images, including image generation \cite{ho2020denoising}, image editing \cite{meng2021sdedit}, and image-to-image translation \cite{li2022vqbb}. Video generation and prediction are effective approaches to understand the real world. Several standard architectures have been utilized in this task, including Generative Adversarial Networks (GANs) \cite{goodfellow2020generative}, flow-based models, auto-regressive models, and Variational Autoencoders (VAEs) \cite{kingma2013auto}. Recently, more diffusion models have been applied in this domain and achieve better video quality and more realistic frames, such as video generation \cite{harvey2022flexible} and text prompt to video generation \cite{villegas2022phenaki}.  

Diffusion models are a class of deep generative models characterized by two main phases: (i) a forward diffusion phase, where the initial data is incrementally disturbed by the addition of Gaussian noise across multiple steps, and (ii) a reverse diffusion phase, where a generative model aims to reconstruct the original data from the noise-added version by progressively learning to invert the diffusion process, step by step. Denoising Diffusion Probabilistic Models (DDPM) represent a common type of generative model designed to learn and generate a specific target probability distribution through a diffusion process. DDPMs have been validated to be more effective than the traditional generation models, such as GANs and VAE.

Generating long videos requires a large amount of computation sources. Some works overcome this challenge with autoregressive based models, such as Phenaki \cite{villegas2022phenaki} and \cite{ge2022long}. However, autoregressive models may lead to unrealistic scene transitions and persistent inconsistencies in extended video sequences because these models lack the opportunity to assimilate patterns from longer footage. To overcome this, MCVD \cite{voleti2022mcvd} employs a training approach that prepares the model for various video generation tasks by independently and randomly masking either all preceding or subsequent frames. Meanwhile, FDM \cite{harvey2022flexible} introduces a framework based on Diffusion Probabilistic Models (DDPMs) that is capable of generating extended video sequences with realistic and coherent scene completion across diverse settings. NUWA-XL \cite{yin2023nuwa} introduces a "Diffusion over Diffusion" architecture designed for generating extended videos through a "coarse-to-fine" method.

In recent years, large language models (LLMs), which are text-based, have seen a surge in popularity \cite{wang2024llmsunderstandsocialnorms}. Additionally, various generative vision-language models (VLMs) have been introduced in the autonomous driving domain. RAG-Driver \cite{yuan2024rag} was proposed to leverage in-context learning for high performance, explainable autonomous driving. We leverage the in-context learning capabilities of EILEV \cite{yu2023efficient} to generate descriptions of driving scenarios. In DriveGenVLM, the in-context VLMs allow us to process videos predicted by the diffusion framework, which can then be recognized by other vision-based models, potentially contributing to decision-making algorithms in autonomous driving. To the best of our knowledge, DriveGenVLM is the first work to integrate a video generation model and a Vision Language Model (VLM) into the autonomous driving domain.

\subsection{Contributions of This Work}

The key contributions of DriveGenVLM are summarized as follows:
\begin{enumerate}
    \item  Apply conditional denoising diffusion probabilistic models to the domain of driving video prediction.
    \item Test the video generation framework in the Waymo open dataset of different camera angles to validate the feasibility for real world driving scenarios.
    \item Utilize in-context vision language model to generate descriptions of the predicted video and validate that these videos can be applied for Vision language model based autonomous driving.
\end{enumerate}
The rest of the paper is organized as follows. Sec.~\ref{sec-preliminory} introduces the preliminary knowledge used in this paper. Sec.~\ref{Solution-approach} illustrates the solution approach. Sec.~\ref{sec:experiments} introduces the setting and results of the experiments. And Sec~.\ref{sec:conclusion} concludes this study.

\section{Preliminory}
\label{sec-preliminory}

\subsection{Denoising Diffusion Probabilistic Models (DDPM)}

The Denoising Diffusion Probabilistic Model is a type of generative model that has gained significant attention in the field of machine learning and computer vision \cite{fu2024genddsgeneratingdiversedriving}. DDPM operates through a forward process that transforms data into noise, and a backward process that reconstructs the original data from the noise. The goal of the forward process is to convert any data into a basic prior distribution, whereas the subsequent objective involves developing transition kernels to undo this conversion. To generate new data points, one begins by drawing a random vector from the prior distribution, then proceeds with ancestral sampling via the reverse Markov chain. The key to this sampling technique is to train the reverse Markov chain to replicate the time-reversed progression of the forward Markov chain accurately.

\begin{figure}[H]
	\centering
	\includegraphics[scale=.38]{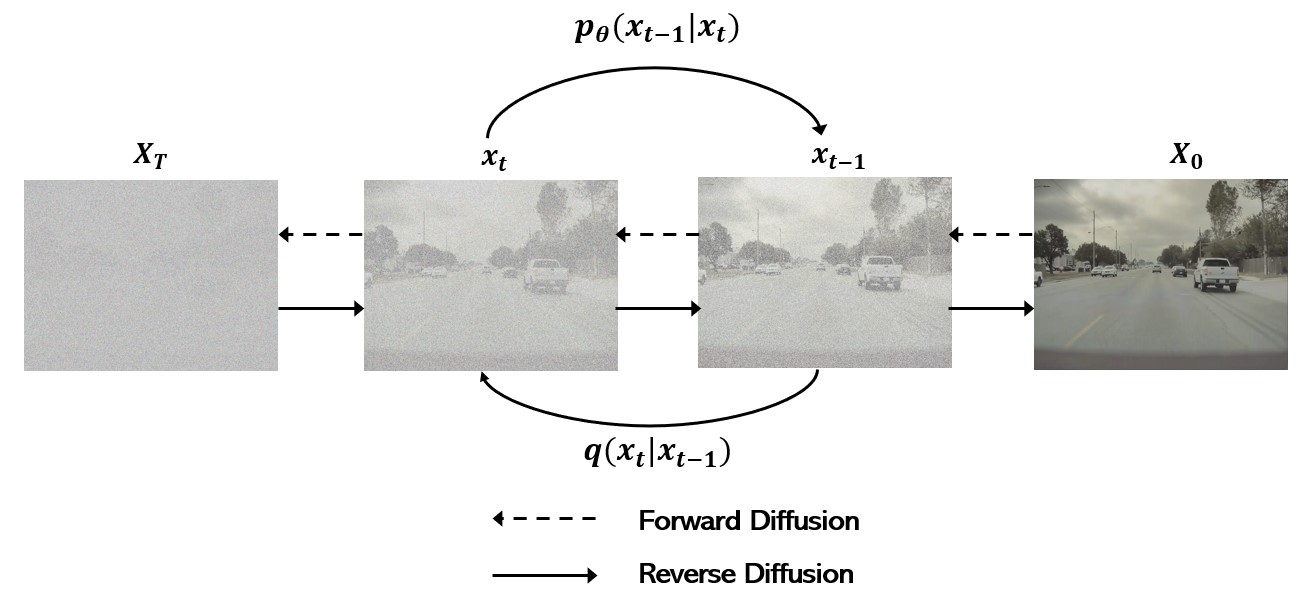}
	\centering 
	\caption{Process of DDPM model.}
	\label{fig:realworld-net}
\end{figure}

For the conditional extension, in which the modeled $x$ is
conditioned on observations$ y$. Given a data distribution $x_0 \sim q(x_0)$, the forward process generates a sequence of random variables $x_1, x_2, ..., x_T$. $x_0$ represents the original, noise-free data, while $x_1$ incorporates a slight amount of noise. This process continues up to $x_T$, which is nearly uncorrelated with $x_0$ and resembles a random sample drawn from a unit Gaussian distribution. The distribution of $x_t$ depends only on $x_{t-1}$, the transition kernel is:

\begin{equation}
\label{eq1}
q(x_t | x_{t-1}) = \mathcal{N}(x_t; \sqrt{\alpha_t} x_{t-1}, (1 - \alpha_t)I).
\end{equation}

The joint distribution is defined by the diffusion process and a data distribution $q(x_0,y)$ in Equ.~\ref{eq2}.
\begin{equation}
\label{eq2}
q (x_{0:T} | y) = q(x_0, y) \prod_{t=1}^T q (x_t | x_{t-1})
\end{equation}

Denoting Diffusion Probabilistic Models (DPMs), these models operate by reversing the diffusion sequence. For given \(x_t\) and \(y\), we use a neural network to estimate \(p_{\theta}(x_{t-1} | x_t, y)\), serving as an approximation for \(q(x_{t-1} | x_t, y)\). This estimation grants us the capability to procure samples of \(x_0\) by commencing with the sampling of \(x_T\) from a standard Gaussian distribution, a choice made due to the diffusion process's initial state resembling a Gaussian distribution. Subsequently, we iteratively sample backwards, from \(x_T\) to \(x_0\), through \(p_{\theta}\). The aggregate distribution of the sampled \(x_{0:T}\) conditional on \(y\) is expressed as:

\begin{equation}
p_{\theta}(x_{0:T} | y) = p(x_T) \prod_{t=1}^{T} p_{\theta}(x_{t-1} | x_t, y)
\end{equation}

Here, \(p(x_T)\) signifies a unit Gaussian distribution independent of \(\theta\). Training a conditional DPM entails the adjustment of \(p_{\theta}(x_{t-1} | x_t, y)\) to closely match \(q(x_{t-1} | x_t, y)\) across the full range of \(t\), \(x_t\), and \(y\) values.
\begin{figure*}[htb]
	\centering
	\includegraphics[scale=0.50]{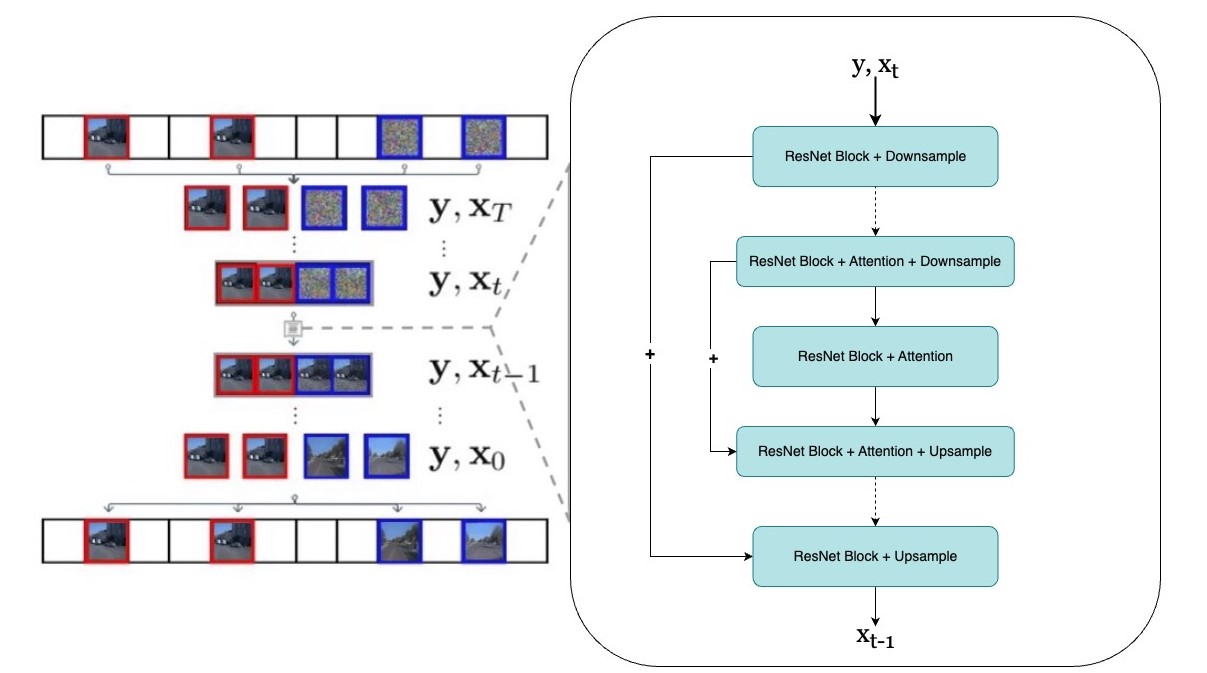}
	\centering 
	\caption{Training Framework Employing U-Net with Diffusion Probabilistic Model (DDPM) Integration.}
	\label{fig:train-arc}
\end{figure*}

\subsection{In-context Learning on VLMs}

\begin{figure}[H]
	\centering
	\includegraphics[scale=.55]{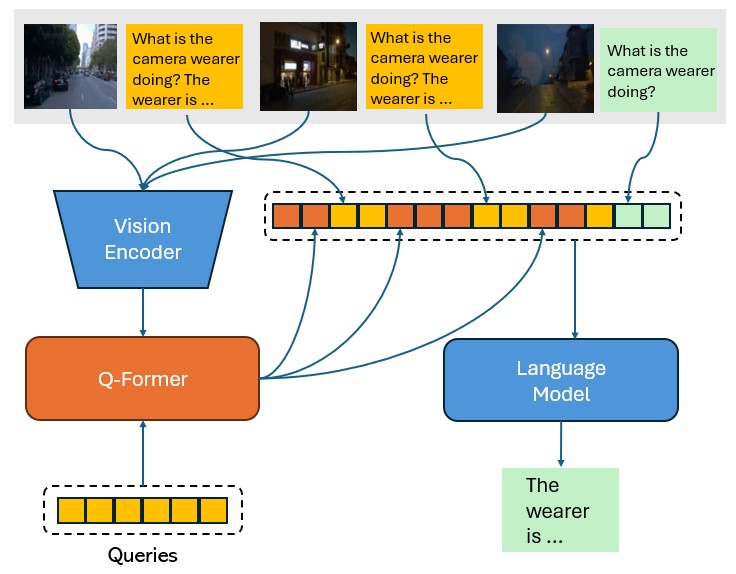}
	\centering 
	\caption{Architecture of EILEV.}
	\label{fig:eilev}
\end{figure}

In-context learning was originally proposed in the paper of GPT-3 \cite{brown2020language}, which refers to the ability of a model to learn or adapt its responses based on the context provided within a single interaction, without any explicit updates or retraining on its underlying model. 

We employ EILEV \cite{yu2023efficient}, a training technique developed to enhance in-context learning in Vision Language Models (VLMs) for first-person videos. As shown in Figure.~\ref{fig:eilev}, EILEV's architecture for an interleaved context-query scenario involves using the unmodified Vision transformer from BLIP-2 \cite{li2023blip} to process video clips. The resulting compressed tokens are mixed with text tokens in the sequence of the initial context-query instance. These combined tokens are then input into BLIP-2’s static language model, which produces new text tokens. This method can generalize out-of-distribution videos and texts and rare actions vis in-context learning. We make use of the pre-trained model to generate language narrations for the driving videos to validate that the generated results are explainable and realistic.

\section{Methodolegy}

Generating long, coherent, and photorealistic videos is still a challenge. The Flexible Diffusion Model (FDM) addresses this issue using a conditional generative model. We adopt a similar approach in DriveGenVLM. To sample coherent videos with a large number of frames, we can sample an arbitrary length of video condition on a small number of frames with a generative model. The goal is to sample coherent photo-realistic videos of driving scenarios with some frames. We utilize a sequential procedure to sample an arbitrarily long video with a generative model that can sample or condition on only a small number of frames at once. 

Broadly, we define a sampling scheme as a series of tuples $\left[\left(X_s, Y_s\right)\right]_{s=1}^S$, where each tuple consists of a vector $X_s$ indicating the indices of frames to be sampled and a vector $Y_s$ showing the indices of frames to use as conditions for the stages $s = 1, \ldots, S$.

\label{Solution-approach}

\subsection{Training Architecture}

We utilize a U-net structure for the DDPM image framework. This architecture is characterized by a sequence of layers that downscale spatial dimensions and then upscale them, interspersed with convolutional residual network blocks and layers that focus on spatial attention. The architecture is illustrated in Figure.~\ref{fig:train-arc}. The DDPM iteratively transforms noise $X_T$ to video frames $X_0$. The boxes with red borders are conditions. The right side shows the UNet architecture of each DDPM step. 

Agorithm.~\ref{alg:sample} illustrates how we sample a video using a sample scheme. The generative model can sample any subsets of the video frames conditioned on other subsets. The model can generate any choice of $X$ and $Y$.

\begin{algorithm}
\caption{Sample a video $v$ given a sampling scheme $\left[(x_s, y_s)\right]_{s=1}^S$. }
\begin{algorithmic}[1]
\Procedure{SampleVideo}{$v; \theta$}
    \For{$s \gets 1, \dots, S$}
        \State $y \gets v[Y_s]$
        \State $x \sim \text{DDPM}(\cdot; y, X_s, Y_s, \theta)$
        \State $v[X_s] \gets x$
    \EndFor
    \State \Return $v$
\EndProcedure

\end{algorithmic}
\label{alg:sample}
\end{algorithm}

\subsection{Sampling Schemes}

\begin{table}[H]
\centering
\begin{tabular}{|c||p{5cm}|}\hline
Sampling Schemes
& Description\\\hline\hline
Autoreg &  Samples $x$ consecutive frames at each stage
conditioned on the previous ten frames.  \\\hline
Hierarchy-2 & Selects the first $x$ frames (large groups) conditioned upon the previous ten frames, then samples consecutive frames within those groups until all frames are sampled.  \\\hline
Adaptive Hierarchy-2 & Selects primary and secondary sampling frames, and adapts based on information gathered during the sampling process to optimize frame diversity using LPIPS distance. \\\hline

\end{tabular}
\caption{Different Sampling Schemes.}
\label{table:sampling}
\end{table}

Each sampling scheme's relative efficacy heavily relies on the dataset at hand, and there is no universally optimal option. In this work, we experimented with three sampling schemes as shown in Table \ref{table:sampling}. The first and the most straightforward scheme adopted is the Autoreg, which samples ten consecutive frames at each step by conditioning on the previous ten frames. Another scheme used was Hierarchy-2 which employs a multi-tiered sampling approach, first tier with ten equidistantly chosen frames covering the unobserved portion of the video, conditioned upon ten observed frames. In the second tier, consecutive frames are sampled in groups, considering the nearest preceding and proceeding frames until all frames are sampled. Lastly, we used Adaptive Hierarchy-2 (Ad), which is only achievable through the implementation of FDM. Adaptive Hierarchy-2 strategically selects conditioning frames during testing to optimize frame diversity, measured by the pairwise LPIPS distance between them.

\section{Experiments}
\label{sec:experiments}

\subsection{Datasets}

The Waymo Open Dataset \cite{sun2020scalability} is a wide-ranging dataset that uses many sensors to aid in the progress of self-driving technology. It contains high-quality sensor data from Waymo's group of autonomous vehicles and is made up of more than 1,000 hours of videos. These videos are taken with various sensors such as LIDARs, radars, and five cameras (front and sides); they give a complete view around the car at all times or what we call 360-degree visibility. This group of data has very careful labeling, including marks for vehicles, people walking, bicycle riders and other things found on the road. This makes it extremely helpful for those working as researchers or engineers in this area to enhance their skills with perception (understanding), prediction (guessing what happens next) and simulation algorithms in self-driving cars. The Dataset V2 format is designed to be usable with Apache Parquet file formats and supported components. Here, a component is a set of related fields/columns that are required to understand each individual field.

\subsection{Experiment Setup}

To validate the algorithm in real-world driving scenarios, we utilize the Waymo Open Dataset, which encompasses diverse real-world environments across several cities. We extracted data for all the five present cameras in the dataset. We then pre-processed the datasets and extract the data from the three cameras being Front, Front-left, Front-right. In total we processed $138$ videos. A total of $108$ videos comprising of all three cameras divided equally were taken for training purposes, while the $10$ videos for each of the three cameras for the test set. The maximum number of frames found for train videos was 199 frames, minimum contained around $175$ frames. So we used $175$ frames as the limit for all videos. The resolution was reduced to $128 \times 128$, and transformed into 4D tensors.

\begin{table}
    \centering
    \renewcommand{\arraystretch}{1.5} 
    \begin{tabular}{|c||c|c|c|}
    \hline
    \textbf{Camera Number} & \textbf{Camera Type} & \textbf{Iterations} & \textbf{GPU Hours}\\
    \hline
    1 & Front & 200,000 & 48 \\
    \hline
    3 & Front-right & 150,000 & 36 \\
    \hline
    2 & Front-left & 100,000 & 24 \\
    \hline
    \end{tabular}
    \caption{Training details for each camera}
    \label{table:train}
\end{table}

The model was operated on an 8-core Intel Cascade Lake processor and an NVIDIA L4 GPU with 24 GB memory in Debian GNU/Linux 11. A batch size of 1 with a learning rate of $0.0001$ was used. The details of each camera training are shown in table \ref{table:train}. The front was trained from scratch without using any pre-trained weights for $200,000$ iterations. Front-right used pre-trained weights from Camera-1 and was trained for $150,000$ iterations, and Front-left used pre-trained weights from Camera-3, trained for $100,000$ iterations. A total of $108$ GPU hours were spent on training.

\subsection{Metrics}

We utilize FVD (Fréchet Video Distance) \cite{unterthiner2018towards}, a metric used to evaluate the quality of videos generated by models in tasks like video generation or future frame prediction. Similar to the Fréchet Inception Distance (FID) used for images, FVD measures the similarity between the distribution of generated videos and real videos. FVD is useful for assessing the temporal coherence and visual quality of videos, making it a valuable tool for benchmarking video synthesis models.

\subsection{Results}

The FVD scores from our experiments on the Waymo Open Dataset for three cameras, which are tested using various sampling schemes, are summarized in Tables \ref{tab:camera1_fvd} \ref{tab:camera2_fvd} \ref{tab:camera3_fvd}. The adaptive hierarchy-2 sampling method outperforms the other two methods.

\begin{table}[h]
    \centering
    \begin{tabular}{lc}
        \toprule
         \textbf{Sampling Scheme} & \textbf{FVD Score} \\
        \midrule
        hierarchy-2 & 1489.0138 \\
        autoreg & 1266.354 \\
        adaptive hierarchy-2 & 1174.563 \\
        \bottomrule
    \end{tabular}
    \caption{FVD Scores for Front Camera.}
    \label{tab:camera1_fvd}
\end{table}

\begin{table}[h]
    \centering
    \begin{tabular}{lc}
        \toprule
        \textbf{Sampling Scheme} & \textbf{FVD Score} \\
        \midrule
        hierarchy-2 & 1295.744 \\
        autoreg & 1401.793 \\
        adaptive hierarchy-2 & 812.425 \\
        \bottomrule
    \end{tabular}
    \caption{FVD Scores for Front-left Camera.}
    \label{tab:camera2_fvd}
\end{table}

\begin{table}[h]
    \centering
    \begin{tabular}{lc}
        \toprule
        \textbf{Sampling Scheme} & \textbf{FVD Score} \\
        \midrule
        hierarchy-2 & 1214.684 \\
        autoreg & 1338.234 \\
        adaptive hierarchy-2 & 1122.159 \\
        \bottomrule
    \end{tabular}
    \caption{FVD Scores for Front-right Camera.}
    \label{tab:camera3_fvd}
\end{table}

\begin{figure}[H]
	\centering
	\includegraphics[scale=.28]{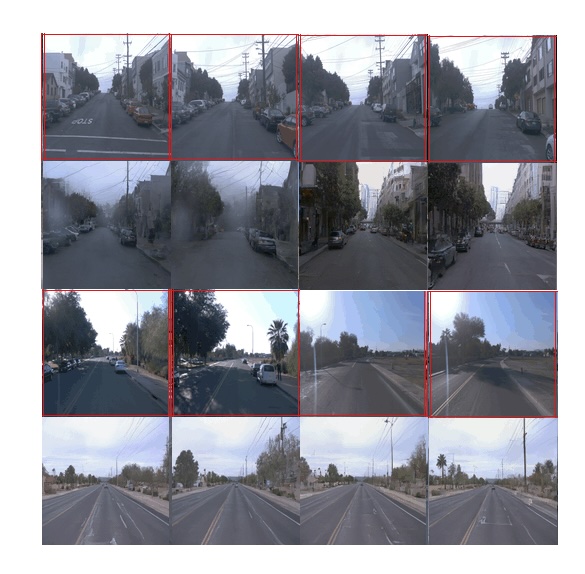}
	\centering 
	\caption{Front Camera - FVD Score: 1174.}
 \label{fig:camera1}
\end{figure}

\begin{figure}[H]
	\centering
	\includegraphics[scale=.23]{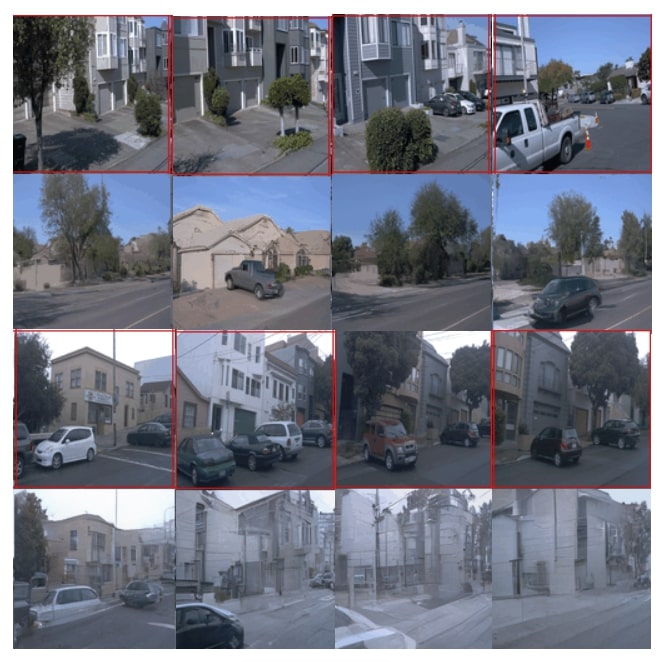}
	\centering 
	\caption{Front-left Camera - FVD Score: 812.}
 \label{fig:camera2}
\end{figure}

\begin{figure}[H]
	\centering
	\includegraphics[scale=.22]{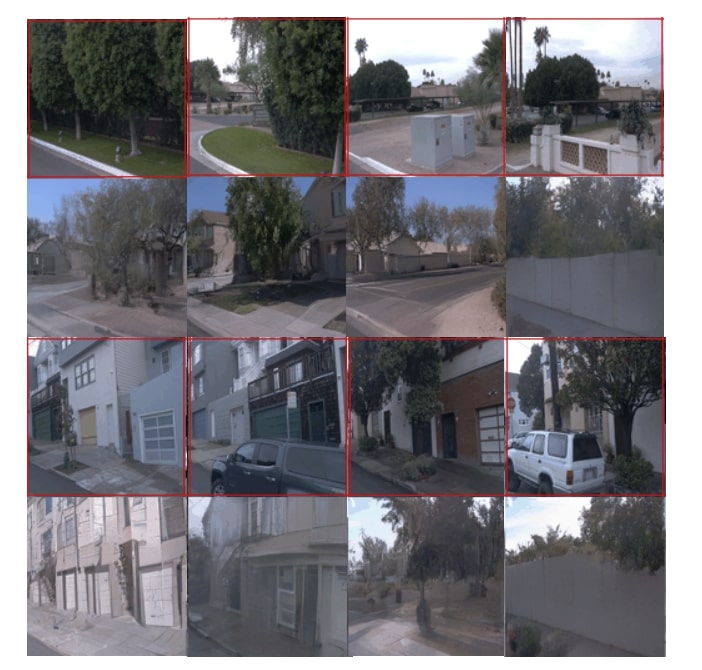}
	\centering 
	\caption{Front-right Camera - FVD Score: 1122.}
 \label{fig:camera3}
\end{figure}

Figure.~\ref{fig:camera1} - \ref{fig:camera3} shows prediction videos generated for each of the three cameras, using the Adaptive Hierarchy-2 sampling schemes yielding the lowest FVD scores.
Each sub-figure contains 2 examples of generation videos for each camera. The frames with red bounding boxes are the ground truth frames, and the predicted frames are below each corresponding frame. The generated videos were conditioned on the first 40 frames for each example. 
The FDM's training on the Waymo dataset showcased its capacity for coherence and photorealism. However, it still struggles with accurately interpreting the complex logic of real-world driving, such as navigating traffic and pedestrians. This limitation is likely due to the additional challenges present in real-world scenarios, which are absent in simulated environments.

\subsection{Prediction Validation by in-context learning.}

To validate that our generated videos are explainable and usable in vision language models, we employ the EILEV pre-trained model on Ego4D, eilev-blip2-opt-2.7b \cite{yu2023efficient} to test our generated driving videos.

\begin{figure*}[htb]
	\centering
	\includegraphics[scale=0.28]{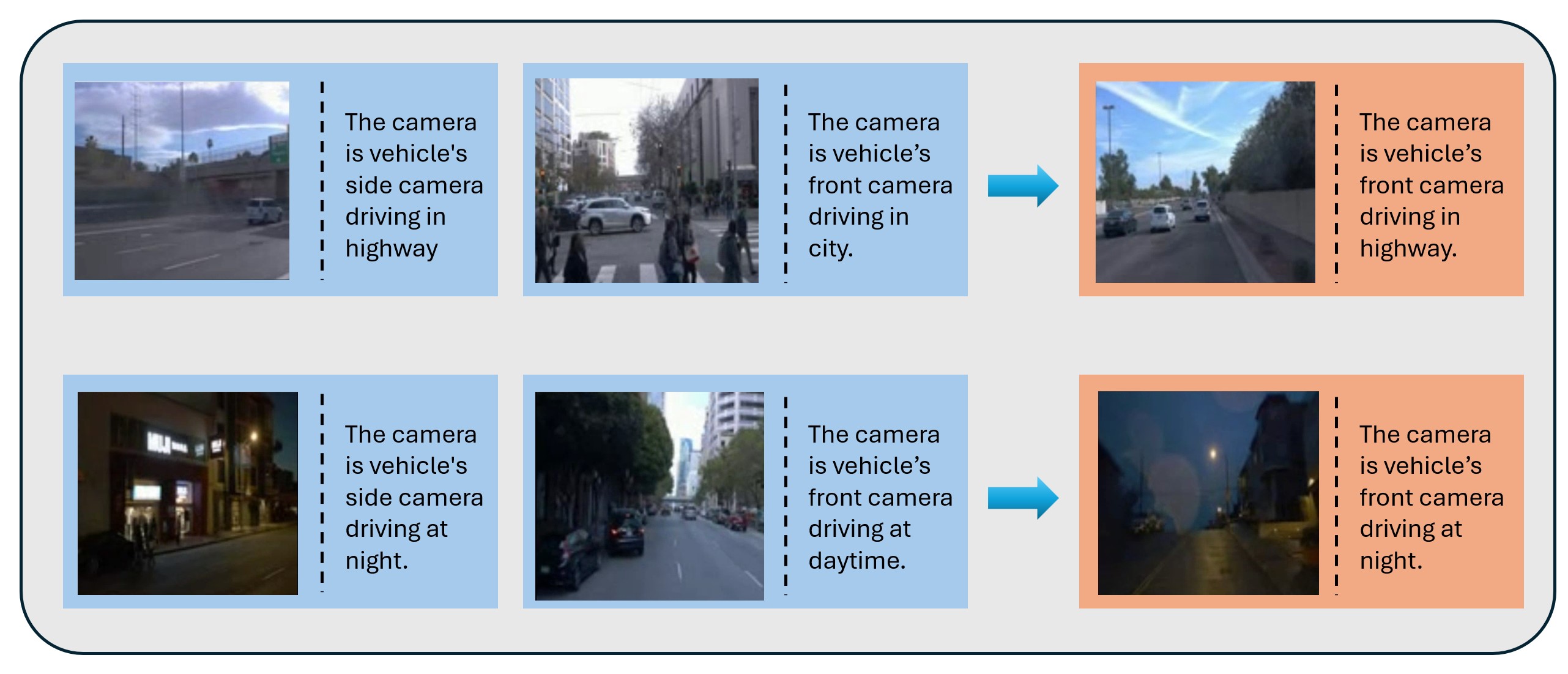}
	\centering 
	\caption{The VLM model with in-context learning capabilities can generate action narrations for unseen driving videos.}
	\label{fig:in-context}
\end{figure*}

We utilize video clips and text pairs that describe the camera angle, driving environment, and time of day. The results are illustrated in Figure.~\ref{fig:in-context}. The action narrations generated by the model are displayed in an orange box. Notably, none of the verb and noun class combinations are shared in the first two videos, as shown in the blue box. As we can observe, the model can identify that the vehicle is driving on a highway with the camera positioned at the front. For the second video, the model recognizes that the vehicle is driving at night with its front camera. The in-context learning pre-trained model on VLMs performs well with the generated model, indicating that the videos are explainable and potentially usable by VLMs-based algorithms.
\section{Conclusions}\label{sec:conclusion}

In summary, training the Denoising Diffusion Probabilistic Model (DDPM) on the Waymo dataset has shown its capability to produce coherent and lifelike images from both front and side cameras. However, it continues to face challenges in accurately capturing the complex dynamics of real-world driving, such as detailing buildings and tracking pedestrian movements. These difficulties are likely due to the complexities inherent in actual driving conditions, which are absent in synthetic datasets.

To explore potential applications of generated videos in Vision-Language Models (VLMs) for autonomous driving, we utilize the pre-trained EILEV model, an in-context VLM, to generate action narrations for the videos. The results indicate that the model can recognize unseen scenarios and generate accurate narrations, demonstrating the potential for deploying VLM-based autonomous driving models that leverage outputs from generative models. The DriveGenVLM framework highlights the potential for using generative models and Vision Language Models (VLMs) together in autonomous driving tasks. For downstream applications, once we obtain narrations of driving scenarios, we can employ large language models to provide guidance to the driver or some language model-based algorithms.

\bibliographystyle{IEEEtran}
\bibliography{ref}

\end{document}